\documentclass[journal,11pt,onecolumn]{IEEEtran}

\usepackage{graphicx}
\usepackage{listings}
\usepackage{graphicx}
\usepackage{url}
\usepackage{float}
\usepackage{algorithm}
\usepackage{subfig}
\usepackage[noend]{algpseudocode}
\usepackage{multirow}
\usepackage{tabu}
\usepackage{diagbox}
\usepackage{longtable}

\usepackage{cite}

\usepackage{amsmath}
\usepackage{array}
\usepackage{fixltx2e}
\begin{document}
%
\title{Sketch2Code: Transformation of Sketches to UI in Real-time Using Deep Neural Network}
%
%
%

\author{Vanita Jain, Piyush Agrawal, Subham Banga, Rishabh Kapoor and Shashwat Gulyani
 \thanks{V. Jain is the Head of Department
of Information Technology, Bharati Vidyapeeth's College of Engineering, Paschim Vihar, New Delhi, 110063 India e-mail: (vanita.jain@bharatividyapeeth.edu)
 
  P. Agarwal, S. Banga, S. Gulyani and R. Kapoor are with the Department of Information Technology, Bharati Vidyapeeth's College of Engineering, Paschim Vihar, New Delhi, 110063 India. (e-mail: me@ipiyush.com, subhambanga26@gmail.com, rishabhkapoor99@gmail.com and  shashwat.gulyani@gmail.com).}
}

\maketitle

\begin{abstract}
User Interface (UI) prototyping is a necessary step in the early stages of application development. Transforming sketches of a Graphical User Interface (UI) into a coded UI application is an uninspired but time-consuming task performed by a UI designer. An automated system that can replace human efforts for straightforward implementation of UI designs will greatly speed up this procedure. The works that propose such a system primarily focus on using UI wireframes as input rather than hand-drawn sketches. In this paper, we put forward a novel approach wherein we employ a Deep Neural Network that is trained on our custom database of such sketches to detect UI elements in the input sketch. Detection of objects in sketches is a peculiar visual recognition task that requires a specific solution that our deep neural network model attempts to provide. The output from the network is a platform-independent UI representation object. The UI representation object is a dictionary of key-value pairs to represent the UI elements recognized along with their properties. This is further consumed by our UI parser which creates code for different platforms. The intrinsic platform-independence allows the model to create a UI prototype for multiple platforms with single training. This two-step approach without the need for two trained models improves over other methods giving time-efficient results (average time: 129 ms) with good accuracy.\\

\end{abstract}

\begin{IEEEkeywords}
Sketches, \and Sketch-to-code, \and User Interface, \and Computer Vision, \and Deep learning, \and UI prototyping,
\end{IEEEkeywords}

%
\IEEEpeerreviewmaketitle

\section{Introduction}
User Interface is the part of an application that defines how a user interacts with the application and enables them to operate the application's capabilities. For easily communicating their ideas or views with the stakeholders, developers and designers have adopted sketching as a medium. Designing UI \cite{Myers:2000:PPF:344949.344959} involves the rapid implementation of UI Sketches as prototypes meant for review. A lot of time is invested by developers to build a prototype for getting a review from stakeholders. The code of the prototype is discarded after this step and the application is built following standard steps of application development. These sketches are then transformed into coded UIs by developers. This is a time-consuming process which prevents them from working on unique features of the application or website.

Generation of code from sketches using machine learning techniques is a relatively new field of research. Understanding Sketches in the form of images by a machine is a problem of Computer Vision since it involves a machine making deductions from sketches, understanding them and extracting logical information from them. Eyiokur et. al. \cite{eyiokur} highlighted this problem and showcased how various combination of methods are required to get good accuracy.

Computer Vision has made unprecedented progress since its beginning. Early research studies in the field of Computer vision were focused on Computer-Aided Design (CAD) as an input \cite{rian2018fractal}\cite{sonmez2018review}. CAD systems produced exceptional results in architectural images manipulation \cite{modi2011architecture}\cite{hablicsek2019algebraic}. Lately, Deep Neural Networks (DNN) have been immensely popular with the introduction of Convolutional Neural Networks (CNN). Object detection in images has been the staple problem in computer vision and CNNs\cite{sezmfl13} have emerged as an optimal solution. Szegedy et. al. \cite{szegedy} exhibited that DNNs can be applied to object detection as well as localizing the objects of various categories precisely. The results showed that simple formulation could yield strong results when used with a 'multi-scale course-to-fine procedure'.

Wang et. al. \cite{wang} used the Convolutional Neural Network for image classification instead of just feature extraction. The image classification task was conducted on ImageNet dataset \cite{ddsllf} and they could further use that system for designing a tracking system. The results showcased how source image when processed through the multilevel convolutional system can help characterize from a different perspective. Chao Ma et. al. \cite{robust} showcased how multiple convolutional layers can be exploited to improve visual tracking. The hierarchical features of convolutional layers can be exploited to represent the original object. Erhan et. al. \cite{erhan} proposed a saliency-inspired \cite{saliency} neural network that could predict class-agnostic bounding box and assign a score to each box which corresponds to the probability of them containing any useful object in them.


Object detection with real-life images has been the prime focus of the majority of the research effort. The models so developed are highly efficient and competent at their job but not so in case the image is a hand made sketch. We've studied various models and looked at how their structure and methodology could or could not be useful for our use case.

Fast R-CNN (Fast Region Based Convolutional Neural Network), proposed by Girshick\cite{g}, is an Object detector that outperformed previously proposed methods of region detection like R-CNN and SPPnets (Spatial pyramid pooling networks) at both speed and reliability. 

Fast R-CNN computes bounding boxes to objects in the image and then runs a classifier on those boxes. The post-processing after classification refines the bounding boxes, eliminates duplicate detection and relocates the boxes. This procedure is slow and hard to optimize as every component needs to be trained separately. This makes Fast R-CNN unsuitable for sketches classification.

In the paper titled “Focal Loss for object detection”, Tsung-Yi Lin et. al. \cite{retnet} presented a novel approach to object detection using one-stage detectors that surpassed the accuracy of the prevalent two-stage detection methods. The groundbreaking accuracy rivalling that of two-stage detectors while being faster was made possible due to what the research team called Focal Loss. The focal loss approach allows training to be focused on a sparse set of difficult examples and removes the set of negatives which may submerge the detector during training.

YOLO\cite{DBLP:journals/corr/RedmonDGF15,DBLP:journals/corr/abs-1804-02767} (You Only Look Once) was also tested in object detection. YOLO functions similarly to a Fully Convolutional Neural Network (FCNN). The architecture splits the input image into a square grid and for each element in the grid, 2 bounding boxes are generated. These bounding boxes are greater than the grid itself and hence consumes more time in processing and training. YOLO learns a generalized representation of objects and it works better in object detection of not so close objects than other methods like R-CNN\cite{gddm} by a good margin. Although YOLO appeared to be a better fit for our model, its practical results were not at par with those of RetinaNet.

Dean et. al. \cite{dean} exploit locality-sensitive hashing with a definite number of hash-table probes which can filter responses in the time-independent if the size of the filter bank.
    
Also, there have been several works published under Automatic Code generation. Based on deep learning, these methods use different neural networks for code generation.

DeepCoder \cite{deepcoder} which presented an approach that utilizes deep neural networks to make predictions about properties of the program that derives outputs from the inputs. DeepCoder was shown to be able to solve simpler problems among competitive programming problem sets and useful as an augmentation for search techniques including enumerative search and Satisfiability Modulo Theories (SMT) based solver. However, most of the methods in DeepCoder are based on Domain Specific Language (DSL) which limits the complexity of the language to be generated and thus narrowing its scope.

The closest related work of generating code from an image was proposed by Beltramelli \cite{b}, a deep learning model that can convert Graphical Screen-shots to code for web, Android and iOS platforms. He compared the task of generating code from a screen-shot to writing a textual description of a scene. His model first described the elements found on screen-shot like buttons, text-bars, labels etc. and then tried to generate a code which is syntactically and semantically correct. This was achieved with the help of CNN, CNN performed an unsupervised feature learning by mapping an input image to a definite size vector. Despite having similar two-step approach adopted in our work, pix2code is dependent on platform based training for the Long Short Term Memory (LSTM) model that it uses in the second step which means it requires retraining for each platform. This separate training also increases the overall time for converting the input into a coded UI application.

As another unique approach to the same problem tackled by pix2code, Nguyen et. al. introduced REMAUI \cite{nc15} which implements the user interface for a mobile application in Android and the iOS environment from screen-shots. Unlike pix2code, REMAUI relies on Optical Character Recognition (OCR) \cite{ocr} coupled with domain heuristics and computer vision techniques such as Canny’s algorithm \cite{canny} and edge dilation \cite{edgedil} for detecting text and edges with considerable accuracy from the input. It finally merges the extracted information and exports it as an application. This approach, while achieving a high pixel similarity between the output application and the original screenshot input, fails to produce code that is similar to how a human developer would implement the UI. Thus, the code produced is not easily understandable and hence difficult for developers to modify or alter it further.

In this paper, a novel approach is proposed through which the prototyping process can be automated to save time on developing UI prototypes. Using this approach, the application components can be developed from its sketches. The system can take picture of a sketch and convert it into its corresponding UI. For this, neural networks are employed and the model is trained on a specially prepared dataset of images of UI sketches. The model identifies the application components and represents it as a JavaScript Object Notation (JSON) structure which is then parsed by a UI parser to create platform specific code. The UI parser can be utilized on any platform and the code will be produced for that platform.

Our first contribution towards making UI prototyping instantaneous, Sketch2Code consists of a Convolutional Neural Network \cite{zhiqiang2017review} that takes the image of a hand-drawn sketch on a plain white surface and creates an Object representation of the UI which is read by the UI parser to generate code for targeted platforms. Our work has the capability to generate UI code for various platforms (i.e., HTML/Web, Android and iOS natives). To enable this, the neural network outputs combined with post-processing produces a UI representation object that is parsed in the next step to produce code for different platforms. The final system supports real-time discussion with automated live reloading as well as a representational UI application with adjustment by human developers and source code deployment.

Rest of this paper is organized in sections below in the following manner. The methodology of the system, Sketch2Code has been discussed in Section \ref{sec:methodology}. Distinct subsections explain the UI detection neural network, the post-processing algorithms and the UI parser. Section \ref{sec:results} deals with the experimental results with a test dataset and presents an analysis of the performance of the model. Section \ref{sec:conclusion} concludes the work done in this paper.

\begin{figure}[ht]
    \centering
    \includegraphics[width=\textwidth]{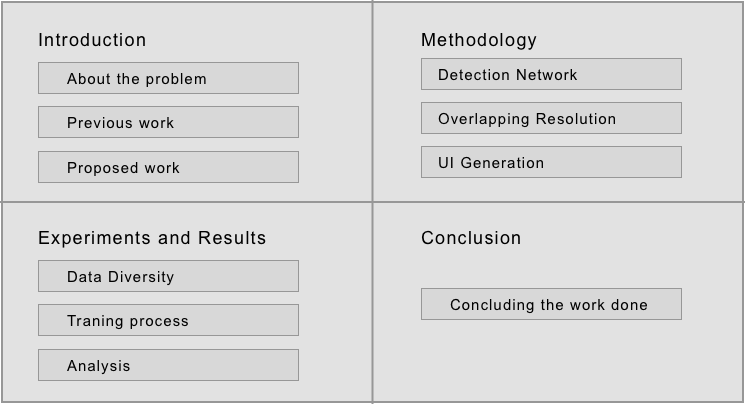}
    \caption{Organization of paper}
    \label{fig:org}
\end{figure}

Figure \ref{fig:org} summarizes the organization of the paper.

\section{Methodology: Sketch2Code} \label{sec:methodology}

\begin{figure}[ht]
    \centering
    \includegraphics[width=\textwidth]{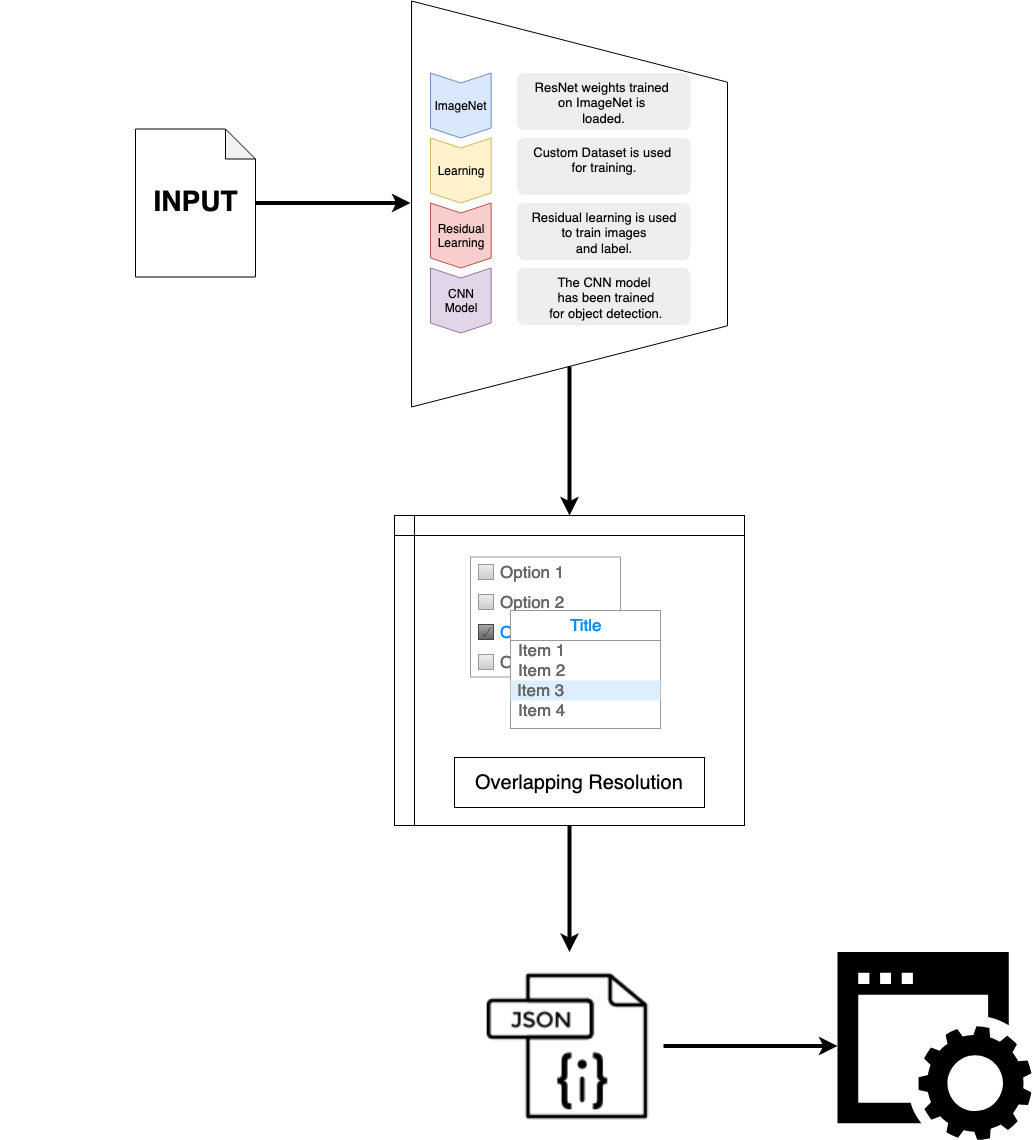}\hfill
    \caption{Representation of the process}
    \label{fig:main}
\end{figure}

The task of generating the corresponding programmed User Interface from a Sketch can be compared to the task of understanding the objects in the image and applying meaningful explanations to them. To do this, the work has been divided into three sub-problems. First sub-problem is a computer vision problem to recognize the objects in images. These objects present are core elements or components of a UI. The object classifier produces bounding boxes with assigned classes for regions where it identifies the objects i.e. UI elements. In certain cases, these bounding boxes overlap each other. This overlapping and how to deal with them constitutes the second sub-problem. The final sub-problem is to create a working prototype application by implementing the UI components identified by computer vision model. Figure \ref{fig:main} represents a schematic overview of the process.


\subsection{UI Detection network}

For the first sub-problem, to understand the context of the elements present in the image, we have employed a deep neural network based object classifier by mapping the input image with a set of classes and generating bounding boxes for the regions where they are present in the image. This mapping allows the training to be done independently of any language or platform restriction. In this way, we have simplified our approach and eliminated the need for Long Short Term Memory (LSTM) \cite{lstm} layers and DSL.


\subsubsection{Input}

\begin{figure}[ht]
  \centering
  \subfloat[Link.]{\includegraphics[width=0.3\textwidth]{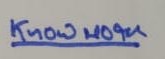}\label{fig:f1}}
  \hfill
  \subfloat[Image.]{\includegraphics[width=0.3\textwidth]{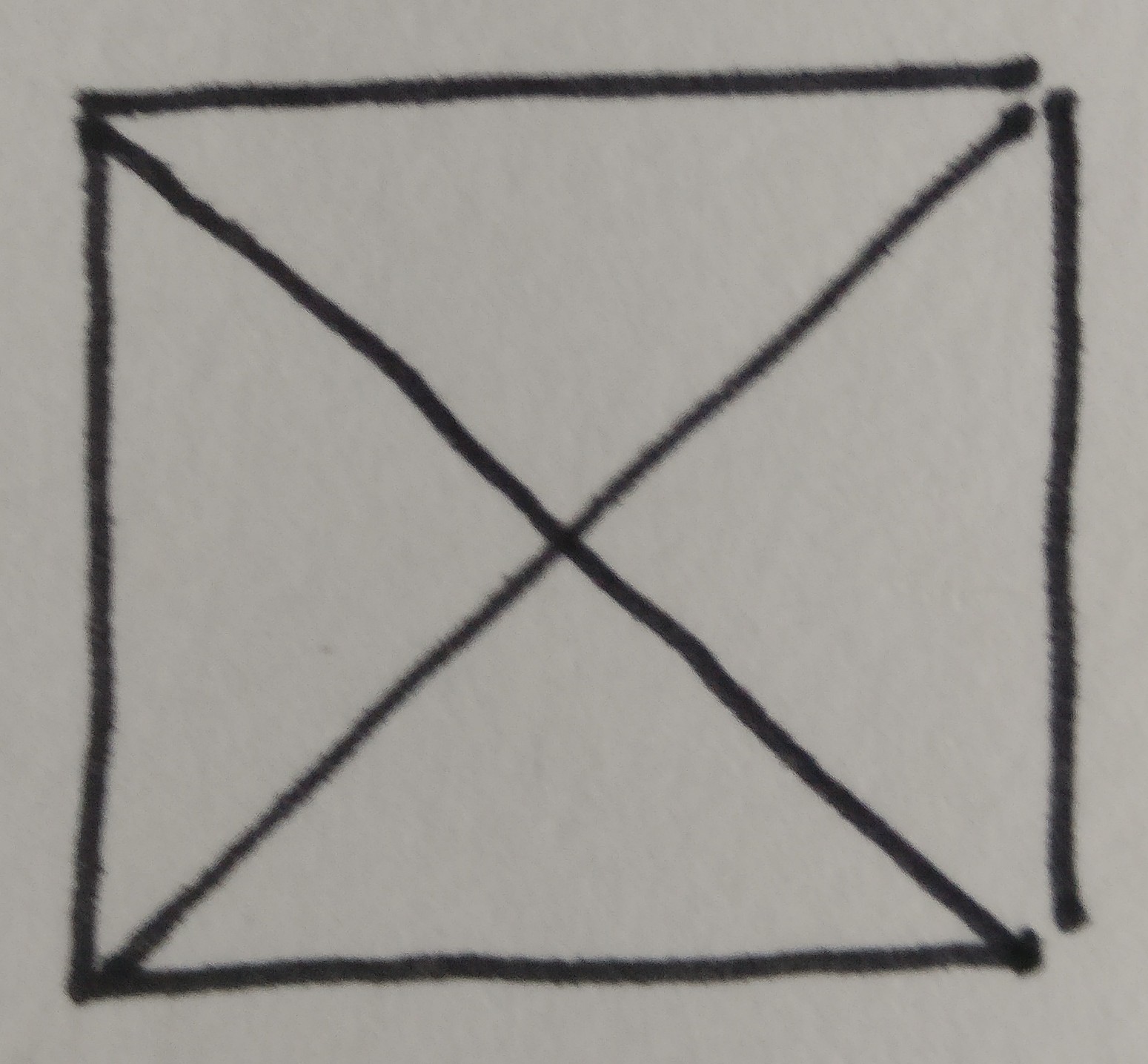}\label{fig:f2}}
  \hfill
  \subfloat[Paragraph.]{\includegraphics[width=0.3\textwidth]{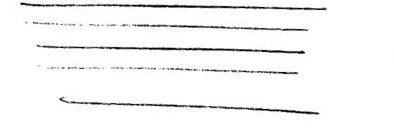}\label{fig:f3}}
  \hfill

  \subfloat[Checkbox.]{\includegraphics[width=0.3\textwidth]{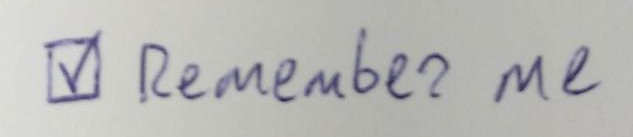}\label{fig:f4}}
  \hfill
  \subfloat[Text box.]{\includegraphics[width=0.3\textwidth]{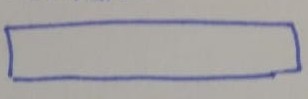}\label{fig:f5}}
  \hfill
  \subfloat[Select box.]{\includegraphics[width=0.3\textwidth]{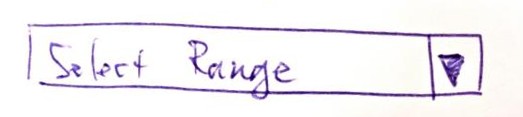}\label{fig:f6}}
  \hfill

  \subfloat[Label.]{\includegraphics[width=0.3\textwidth]{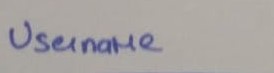}\label{fig:f7}}
  \hfill
  \subfloat[Heading.]{\includegraphics[width=0.3\textwidth]{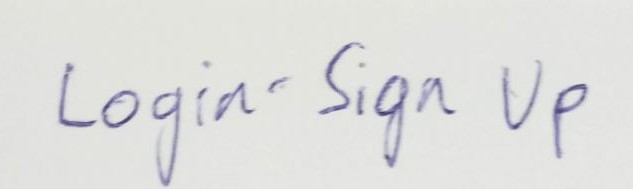}\label{fig:f8}}
  \hfill
  \subfloat[Radio input.]{\includegraphics[width=0.3\textwidth]{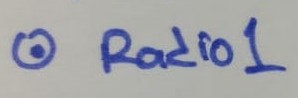}\label{fig:f9}}
  \hfill

  \subfloat[Button.]{\includegraphics[width=0.3\textwidth]{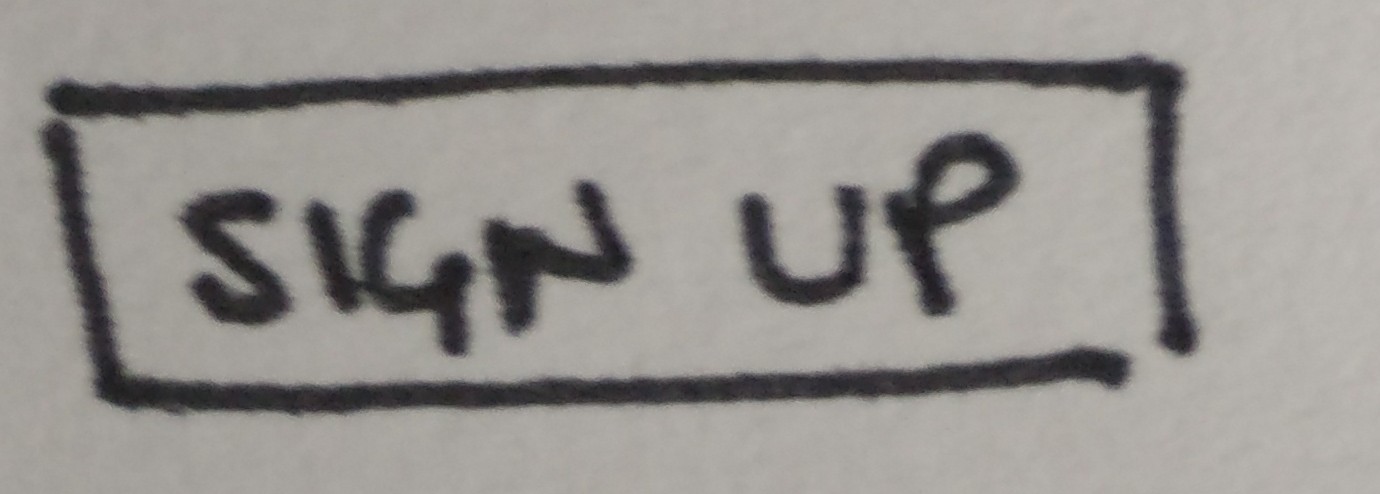}\label{fig:f10}}
  \caption{Sample Sketch of all ten classes of Dataset.}
  \label{Figure:classes}
\end{figure}

The dataset for the input is aimed at creating several instances of sketches of fundamental components in user interfaces. These sketches are drawn by different individuals by following conventions of user interface sketching. The end result is a collection of sketches containing specimens of the 10 most used classes of UI components. On surveying many websites and UI of various applications in domains of the web, Android and iOS, we found that these are the most used classes and hence we trained our model on these classes. The sample of these classes is shown in Figure \ref{Figure:classes}.

These input images can be of any dimensions and are resized to a specific size before being fed into the input layer of the neural network. In our experiments (detailed in a later section) the images are resized until the lesser and larger of the two sides of the image is equal to 800 and 1333 pixels respectively. An input image is a sketch consisting of one or more classes on a white background.

\subsubsection{Neural Network}


Our network is built upon the RetinaNet Object detection architecture as it is a fast, time-efficient method for Single Stage Detection (SSD). The network uses the 50-layer variant of ResNet blocks. Larger ResNet variants were not used as they increased the complexity of the model without enhancing performance. This can be attributed to a limited number of classes and training examples which means larger models merely added redundant parameters.

RetinaNet detector improves in accuracy over other detection models due to its novel loss function which is called focal loss. The focal loss is used to make the training concentrated on difficult to classify examples. We make use of Feature Pyramid Networks (FPN) \cite{fpn} in the architecture to obtain multiple scale information and achieve classification and box regression in one stage.

\begin{figure}[ht]
    \centering
    \includegraphics[width=\textwidth, height=2.5in]{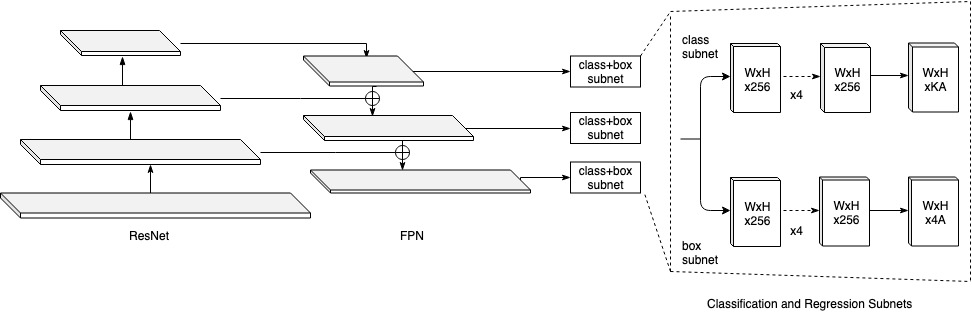}\hfill
    \caption{Neural network architecture}
    \label{fig:retinanet}
\end{figure}

The complete architecture of the neural network is described in Figure \ref{fig:retinanet}.

\noindent \textbf{ResNet:} The neural network utilizes 50 layer variants of ResNet blocks trained on ImageNet dataset. The ResNet layers are used for residual training and perform transfer learning pertaining to our input dataset consisting of ten classes. Due to the fewer number of classes, other variants consisting of a higher number of ResNet layers were not useful.

\noindent \textbf{Feature Pyramid Network:} The network also consists of Feature Pyramid Network for multi-scale feature maps that contain better quality information than the regular feature pyramid for object detection. FPN uses a bottom-up and a top-down pathway along with lateral connection with 1 X 1 convolution filter. The bottom-up pathway is constructed by decreasing the spatial resolution while the semantic value goes up. The top-down pathway is constructed by utilizing the higher semantic features and imbibing them in higher resolution layer. The lateral connections between the top-down reconstructed layer and feature map are needed to predict the location better since the downsampling and upsampling affects the precision of object detection.
\newpage
Two task-specific subnets are joined at each layer of the FPN top-down reconstructed pathway. They are described below:

\noindent \textbf{Classification subnet:}
This subnet is used to predict the probability of the presence of an object at every spatial position at every anchor and object class. At every Feature Pyramid Network (FPN) level a fully convolutional network is attached. The parameters are shared across at all pyramid levels. The subnet only uses a $3\times3$ convolutional filter and does not share the parameters with Box Regression Subnet discussed in the next section.

\noindent \textbf{Box regression subnet:}
Every feature map is tagged with a regression subnet which is parallel to classification subnet. The design of regression subnet and classification subnet are similar. However, the last layer is $3\times3$ with $4X$ linear outputs per spacial locations i.e. $X$ anchors per spatial location. 4 outputs predict relative offset between anchor box and ground-truth box.

\noindent \textbf{Focal Loss function:}
During the training phase of the neural network, the training algorithm tries to find the most appropriate set of weights and biases to fit the training set. A loss function is defined that calculates the loss for the current prediction output and the expected value for an instance of a training set. This choice of this function affects the training as it is used in the backpropagation algorithm that adjusts the weights and biases of the neural network. Often times, a huge class imbalance between foreground and background classes can submerge the detector which makes the training inefficient and produces inaccurate results for hard examples. Lin et. al. proposed to solve this problem naturally by incorporating the balancing factor in the loss function itself. This is what they called focal loss. The exact equation of the balancing factor was found to be insignificant as different forms of the focal loss gave similar results.

The focal loss function we used was the same as the one used by Lin et. al. in their experiments.
\\
The following is the formulae for the Cross-Entropy (CE) loss (see equation \ref{eqn:CE}).

\begin{equation}
CE (x,z) = 
\left\{
	\begin{array}{ll}
		-log(x)  & \mbox{if } z = 0 \\
		-log(1-x) & \mbox{otherwise}
	\end{array}
\right.
\label{eqn:CE}
\end{equation}

\begin{tabbing}
    \hspace{12em} where $x \in \{0,1\}$ is model estimated probability,\\
    \hspace{12em} and $z \in \{\pm1\}$ specifies ground-truth class. 
 \end{tabbing}

As seen in the above formulae, the cross-entropy loss function is affected by the distribution among correctly classified examples and the presence of incorrect classification. Balanced cross entropy function takes into account the importance of positive and negative examples. This function is presented in equation \ref{eqn:BCE}.

\begin{equation}
CE (x,z) = 
\left\{
	\begin{array}{ll}
		-\alpha  log(x)  & \mbox{if } z = 0 \\
		-\alpha  log(1-x) & \mbox{otherwise}
	\end{array}
\right.
\label{eqn:BCE}
\end{equation}

\begin{tabbing}
    \hspace{12em} where $\alpha \in \{0,1\}$ is weighting factor,\\
 \end{tabbing}

Introduction of focal loss instead of a conventional cross entropy function is done to make the function more sensitive to hard to classify examples. This approach makes training more efficient since the balanced cross entropy function misses the differentiation between easy and hard examples. Given below is the form of focal loss function used in the experiments (equation \ref{eqn:FL}).

\begin{equation}
    FL(x_t) = -\alpha (1 - x_t)^\gamma log(x_t)
    \label{eqn:FL}
\end{equation} 
\begin{tabbing}
    \hspace{14em} where $(1 - x_t)^\gamma$ is modulating factor,\\
    \hspace{14em} and $\gamma \geq 0 $ is tunable focussing factor. 
 \end{tabbing}

We have used the $\alpha$-balanced from since it performs better than the non-$\alpha$-balanced version of the focal loss function.

\subsubsection{Output}
\begin{figure}[H]
    \centering
    \includegraphics[height=3in]{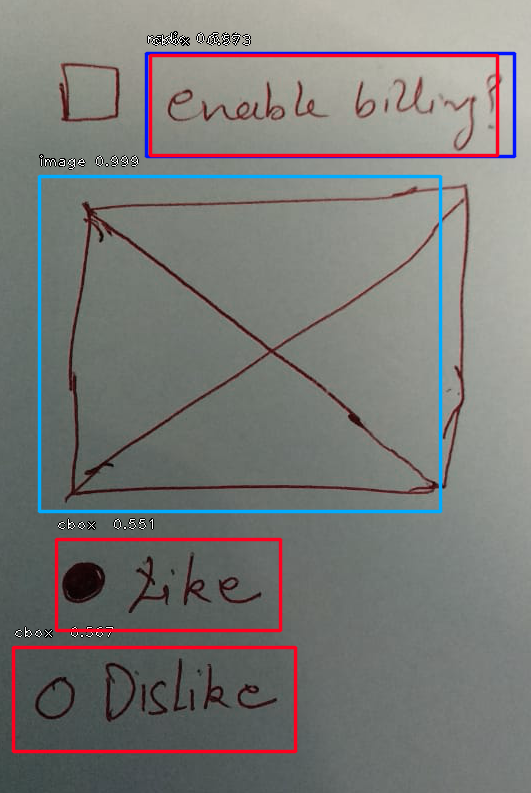}
    \caption{An example of output from the network}
    \label{Figure:CNNOutput}
\end{figure}

The output from the network is a list of coordinate sets for bounding boxes surrounding the UI elements on the sketch with class predictions for them. The output from the network has been represented by drawing the bounding boxes on the original image and shown in Figure \ref{Figure:CNNOutput}.

\subsection{Overlapping Resolution} \label{overlap}

The object detection model is trained to detect and assign classes to regions where a UI component is recognized. It provides a confidence score for each box with an assigned class for the UI component contained in the box. Sometimes these boxes with different classes may overlap each other with varying degrees. This is the second sub-problem we need to resolve. This overlapping of different classified boxes can be the result of two types of situations:\newline
1. Two components in close proximity\newline
2. A component recognized as two different components

The first case happens often with components which generally appear together and close to each other. For example, a button and a text-box in a UI containing a web form. Testing and analysis of such instances revealed that these cases are often representative of more than just closeness of the components. These are often cases of certain groups of elements that appear together which are either tied to each other through some properties or make a composite UI element. Continuing with the button and text-box example, these components when appearing close to each imply that the button is tied to that text-box as part of the same form. In terms of user experience, the behaviour is represented in form validation which means that the button is disabled (greyed-out) when the validation on text-box fails.

The latter case is a false positive by our object detection model. In the absence of a human understanding of UI elements, it misreads, for example, a check-box with a title as a check-box and a label. This is exemplified by the output in Figure \ref{Figure:CNNOutput}. This kind of overlap is quite large and signifies that only one of these multiple classes should be retained. Thus, it needs a filtering algorithm to adjudge the correct class based on the current classes assigned to roughly the same region.

Investigating these cases in juxtaposition, we postulated that an overlap of less than 50\% means that the case belongs to the first category and if it is 50\% or above then it belongs to the second category.

Following equation calculates the area of the overlapping area.

\begin{equation}
    O_{a}(B_{1}, B_{2}) = d_{x} * d_{y}
\end{equation}

Here $O_{a}$ calculates the area of the overlapped region using $B_{1}$ and $B_{2}$, the bounding boxes for the overlapped components. Here, $d_{x}$ is the overlapping length of the boxes along the x-axis and similarly, $d_{y}$ is along the y-axis. 

\begin{equation}
    d_{x} = |B2.x - B1.x|
\end{equation}

\begin{equation}
    d_{y} = |B2.y - B1.y|
\end{equation}

To handle the first case, we've developed an algorithm (see algorithm \ref{alg:filterOverlap}) that takes in the overlapping box's class type and a dictionary identifying components that are likely to be interrelated through properties or parent components. If any such interaction is possible, it represents that in the UI representation through properties of the components or through parent components containing the components with overlapping boxes as its children.

\begin{algorithm}
\caption{Overlapping Classes Filtering Algorithm}\label{alg:filterOverlap}
\begin{algorithmic}[1]
\Procedure{OverlapFilter}{$ele,ele2$}

\State $a\gets type(ele1)$
\State $b\gets type(ele2)$
\If {$matchConflict(ele1, ele2)$}
    \If {$priority(a) < priority(b)$}
        \State \textbf{return} $ele1$\Comment{Element 1 will be kept}
    \Else
        \State \textbf{return} $ele2$\Comment{Element 2 will be kept}
    \EndIf
\EndIf

\EndProcedure
\end{algorithmic}
\end{algorithm}

\begin{figure}[ht]
    \centering
    \includegraphics[width=\textwidth]{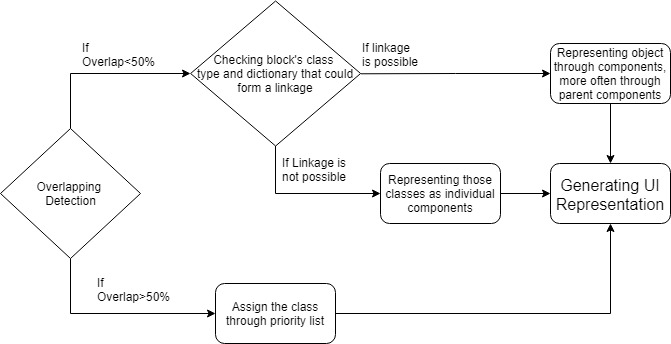}\hfill
    \caption{Overlap Detection process}
    \label{Figure:overlap}
\end{figure}

The whole process has been summarized in Figure \ref{Figure:overlap}.

\subsection{UI Parser} \label{parser}

On being given an image containing a sketch of UI components on the white surface, the trained model will give an output representation of what it recognizes in the image. This representation will be used by the UI parser to create the resulting application. The UI representation structure is an object containing the types of the identified components of UI and their properties.

A web UI application is rendered when the browser loads its Document Object Model (DOM) in its memory. Many frameworks use the concept of Virtual DOMs to manage modification to the UI based on user input that reflects on the browser when the Virtual DOM is synced with the "real" DOM. The object notation used for UI representation is not much different from that of virtual DOM. Furthermore, if the application being built utilizes a virtual DOM, the object notation used for UI representation resembles the initial structure of virtual DOM thus such frameworks can make use of UI representation object generated by the model in building UI. However, the resulting application could be built using any technology stack, it maybe any type of UI application - a browser single page application (SPA), a traditional web page or a mobile application. The presented work can facilitate the creation of UI components of an application using any technology stack and targeted for any platform. The UI representation object generated by the model only needs to be read by a parser supporting creation of UI components for the targeted platform and technology framework. 

The UI parser is the final step where the UI elements present in the UI representation object are transformed into the coded application which can be executed on the target platform. For this work, we deployed a UI parser based on react UI library consuming a NodeJS server, a WebSocket\cite{websocket} connection between the client and the server. The generated file is an HTML document containing the UI components which can be run on any browser.

\section{Experiments and Results} \label{sec:results}

The dataset we have prepared is aimed at creating a collection of fundamental building boxes of a user interface. This is done by creating images consisting of 10 different classes of UI components. These classes are Checkbox, Textbox, Button, Label, Radio Buttons, Heading, Combobox, Link, Image and Paragraph. 

\begin{figure}[ht]
    \centering
     \subfloat{\includegraphics[width=0.48\textwidth]{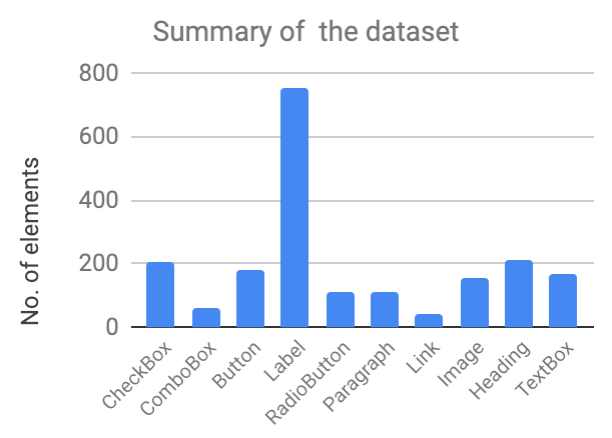}}
     \hfill
     \subfloat{\includegraphics[width=0.48\textwidth]{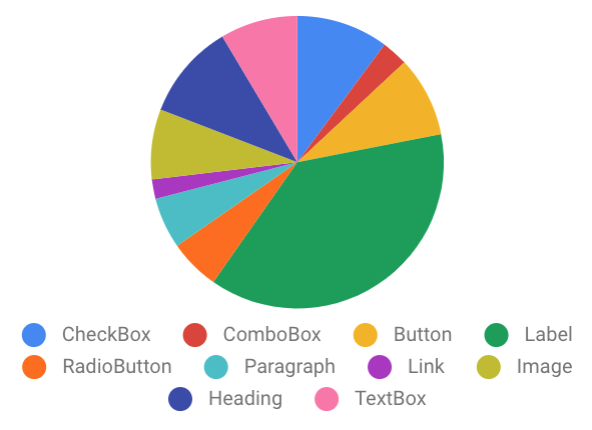}}
     \hfill
    \caption{An overview of training data}
    \label{fig:trainingdata}
\end{figure}

The sketch images are of varying dimensions which are re-sized during pre-processing of the dataset. A collection of 149 sketches, containing 2001 samples of elements, is used for the training process. A summarized view of the collection of the dataset is represented in Figure \ref{fig:trainingdata}.

\begin{figure}[ht]
  \centering
  \subfloat[Low lighting]{\includegraphics[width=0.23\textwidth]{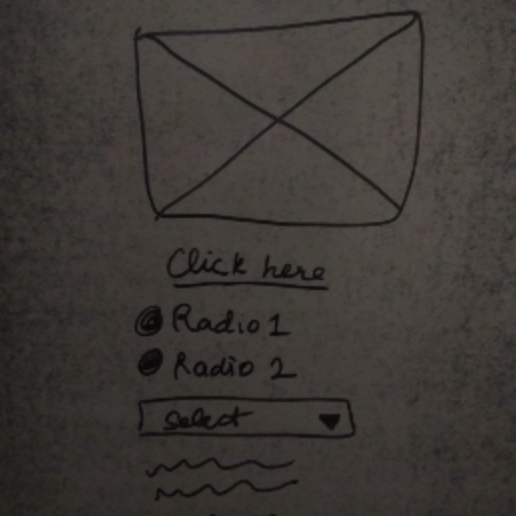}\label{diverse:f1}}
  \hfill
  \subfloat[Standard lighting]{\includegraphics[width=0.23\textwidth]{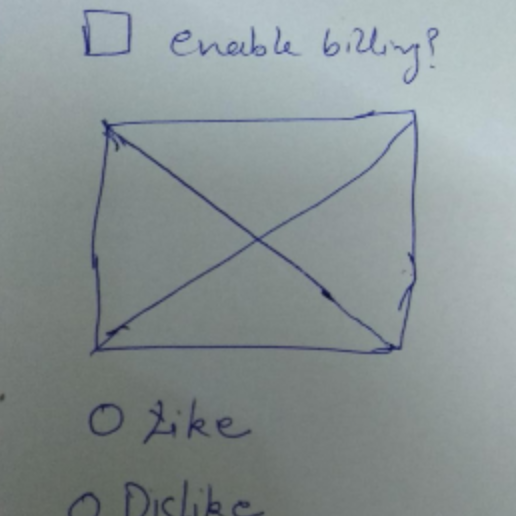}\label{diverse:f2}}
  \hfill
  \subfloat[Standard with Black ink]{\includegraphics[width=0.23\textwidth]{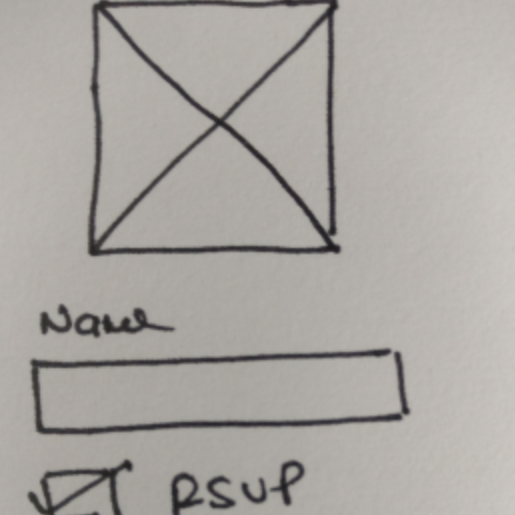}\label{diverse:f3}}
  \hfill
  \subfloat[Whiteboard sketch]{\includegraphics[width=0.23\textwidth]{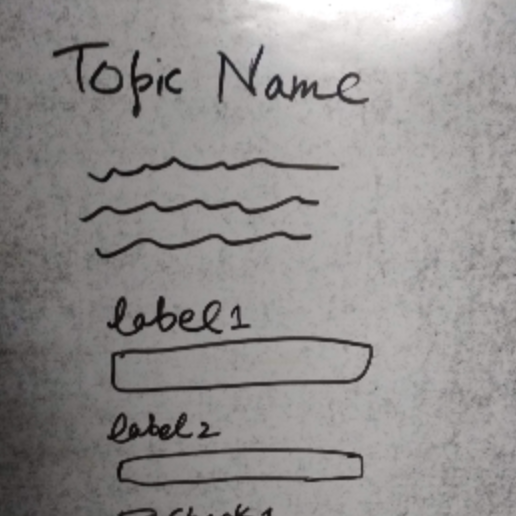}\label{diverse:f4}}
  \caption{Diverse images for our training set}
  \label{diversity}
\end{figure}

One of the key features of the dataset is the inclusion of images with varying levels of lighting. This allows the network to adjust to these lighting conditions, shadow effects and high saturation regions in images. This diversity in dataset simulates real-life situations in which a user may take an image. The overall effect is that the efficiency in handling various cases is improved. For instance, in Figure \ref{diversity}, (a), (b), (c) and (d) show different lighting conditions and background that have been considered for preparing the dataset.

\subsection{Pre-Processing} 
The CNN model is provided with two different comma-separated value (CSV) files: the first one is for annotations, and the second one is for mapping classes with an Integer Identifier (ID).
\newline The annotation file contains six different values: \textit{file-location, x-min, y-min, x-max, y-max, class}; where \textit{x-min, y-min, x-max, and y-max} are dimensions of the bounding box of the annotated UI element in a particular image. This file contains a separate record for each UI element in the image. The classes file contains a list which maps an integer to classes present in annotations file. The network is fed with integer values of the classes for training and prediction.

\subsection{Training Process} 
Before training, the network is fed with Resnet50\cite{hzrs} weights which are trained on ImageNet\cite{ksh12} dataset. Thereafter, the training starts as the training dataset is passed to the network.

\begin{figure}[ht]
    \centering
    \includegraphics[height=3in, width=0.7\textwidth]{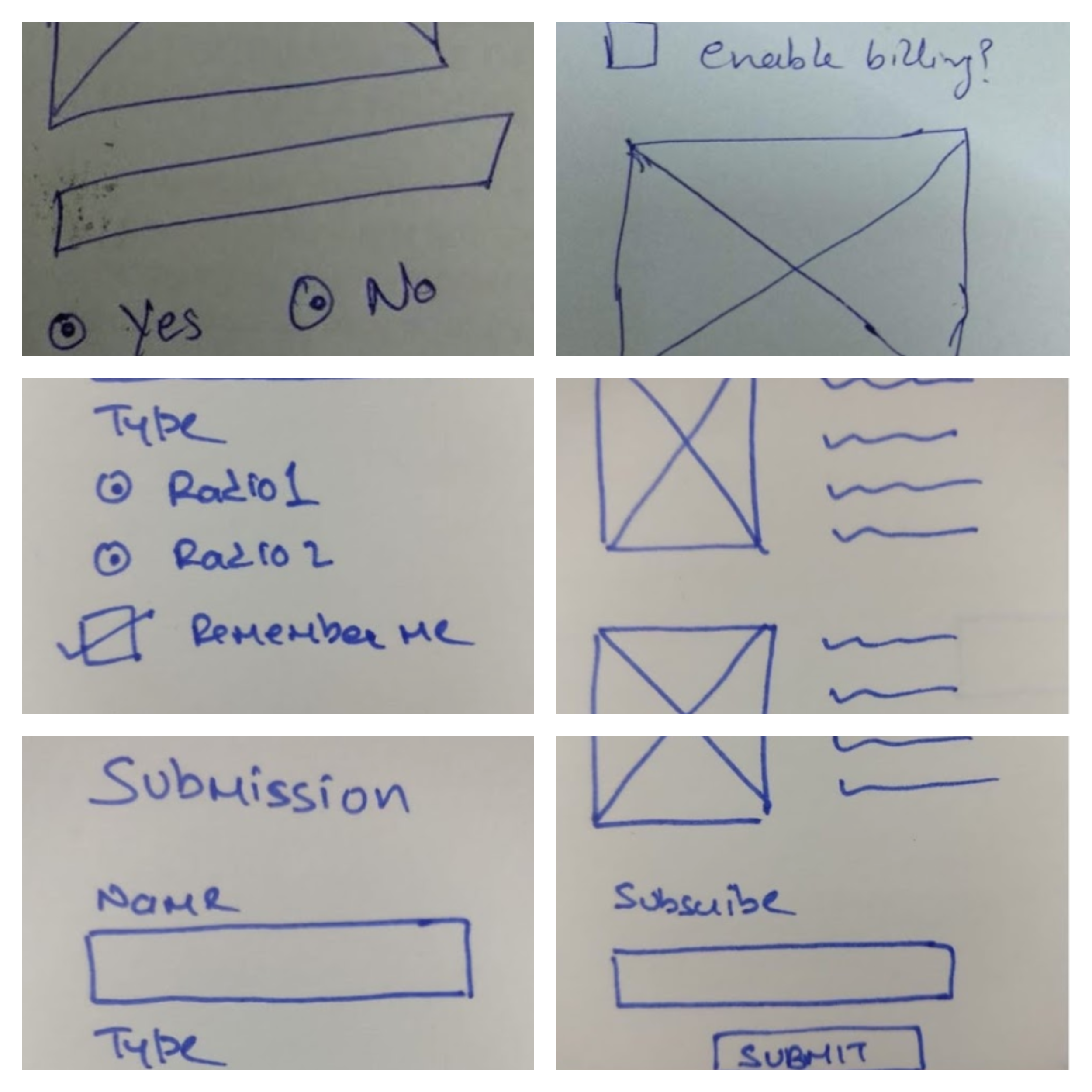}\hfill
    \caption{Sample of Training Dataset}
    \label{Figure:trainingsample}
\end{figure}

Training process has been done to 50 epochs with each epoch consisting of 1000 batches of components in training images. A total of 10 classes and a separate class as the background has been trained, where the white background has been taken for all images. These sketches are drawn both on paper and whiteboard. The training set has been illustrated in Figure \ref{Figure:trainingsample}.\newline\newline

\newpage

\begin{longtable}[c]{| c | c |}

 \caption{Test Dataset.\label{table:testset}}\\
 
 \hline
  Sketch Elements   & Occurrences\\
 \hline
 \endfirsthead
 
 \endhead
 
 \hline
 \endfoot
 
 \hline
 \endlastfoot
\small 
 Heading & 17\\
 CheckBox & 34\\
 Radio & 28\\
 SelectBox & 12\\
 Label & 29\\
 Link & 20\\
 Button & 19\\
 Image & 22\\
 Para & 10\\
 TextBox & 29\\
 \end{longtable}
 
\noindent\textbf{Test set}: The developed model has been tested using 50 distinct sketches (apart from the ones used for training) consisting of different components combinations. The number of instances of each class in the test set has been shown in Table \ref{table:testset}.

\subsection{Analysis}
\begin{figure}[ht]
    \centering
    \includegraphics[height=2in]{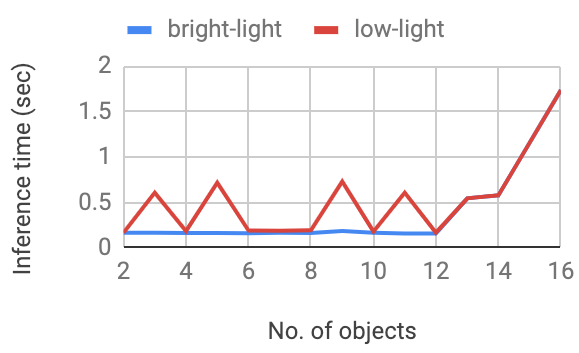}\hfill
    \caption{Performance of the model}
    \label{performance}
\end{figure}

The inference time of our CNN model is shown in Figure. \ref{performance}. This graph shows the inference time for finding elements from a sketch image. The inference time increases as the number of elements in a page increases. Moreover, these results are obtained by training only 2001 components drawn on 149 images. We were able to get this remarkable result despite a relatively small training dataset. Figure \ref{performance} and further analysis suggests that improved performance and higher gains can be achieved if training dataset is increased by adding and labelling more examples of elements drawn by different people.


Two different conditions, bright-light and low-light, were used to prepare the test set. The graph in Fig \ref{performance} represents the inference time for finding elements from the sketches with shadow as a major factor. In the case of images taken at low-light conditions, pixels are darker, thereby increasing the inference time for identifying elements. This is mainly due to difficulties in identifying the background class and to distinguish it from other classes. On the other hand, bright-light condition causes shadow regions when there is an obstruction in light's path as well as over-saturation on reflective backgrounds. Shadow causes the region to behave similarly to low light conditions and increases the inference time. Over-saturation makes the inference difficult resulting in inaccuracies.

\subsubsection{Ablative Analysis}
We performed inference through the network on a set of images with diverse light conditions as shown in Figure \ref{diversity}. The images in this set are re-sized to the same dimensions and are fed into the network and results are analyzed to see how lighting contributes to it.

\begin{table}[ht]
    \centering
    \subfloat[For t = 0.126s]{
    \begin{tabular}{|c|c|c|}
    \hline
          & Predicted Positive & Predicted Negative\\
    \hline    
       Actual Positive & 2 & 2 \\
    \hline
        Actual Negative & 2 & 0 \\
    \hline
    \end{tabular}}
    \bigskip

    \subfloat[For t = 1.76s]{
    \begin{tabular}{|c|c|c|}
    \hline
          & Predicted Positive & Predicted Negative\\
    \hline    
        Actual Positive & 2 & 1 \\
    \hline
        Actual Negative & 1 &  0\\
    \hline
    \end{tabular}}
    
    \bigskip
    \subfloat[For t = 0.1296s]{
    \begin{tabular}{|c|c|c|}
    \hline
          & Predicted Positive & Predicted Negative\\
    \hline    
        Actual Positive & 3 & 0 \\
    \hline
        Actual Negative & 0 &  0\\
    \hline
    \end{tabular}}

    \bigskip
    \subfloat[For t = 0.1298s]{
    \begin{tabular}{|c|c|c|}
    \hline
          & Predicted Positive & Predicted Negative\\
    \hline    
        Actual Positive & 2 & 4 \\
    \hline
        Actual Negative & 0 &  0\\
    \hline
    \end{tabular}}
    \caption{Results on a diverse set of images for different Inference Times}
    \label{tab:diverse}
\end{table}
Table \ref{tab:diverse} shows that the maximum time is taken in case (b) which is a standard case with shadow is present over the elements in the image. This illustrates that a shadow directly affects the inference time for the model. Also, in the case of (a) which is an image with very low light, the inference time may be small but the prediction is inaccurate. Further, good lighting conditions with a proper white background in (c) give high accuracy along with low inference time. In (d), bad lighting (over-saturated image with a reflection) and an inconsistent background which results in low accuracy.

\subsubsection{GPU Analysis}
The results in Figure \ref{performance} are generated on a NVIDIA GeForce GTX 1060 6GB GPU with cuDNN (CUDA Deep Neural Network Library) acceleration. The processor used is Intel i5 2500K and Ubuntu 18.04 operating system.  The inference time for any specific image is recorded by excluding the time to load the image into the memory.

\subsection{Outputs}
To initiate the conversion of a sketch into corresponding coded UI, an image of UI sketch is passed to the trained network. Our network identifies the objects as elements of UI in the image and provides the bounding boxes with classes on them. These bounding boxes with classes are then passed to Overlapping detection procedures as described in section \ref{overlap} and then the UI representation object is created. In the end, this representation is passed to UI parser as explained in section \ref{parser}.

\begin{figure}[ht]
    \centering
    \includegraphics[height=4in]{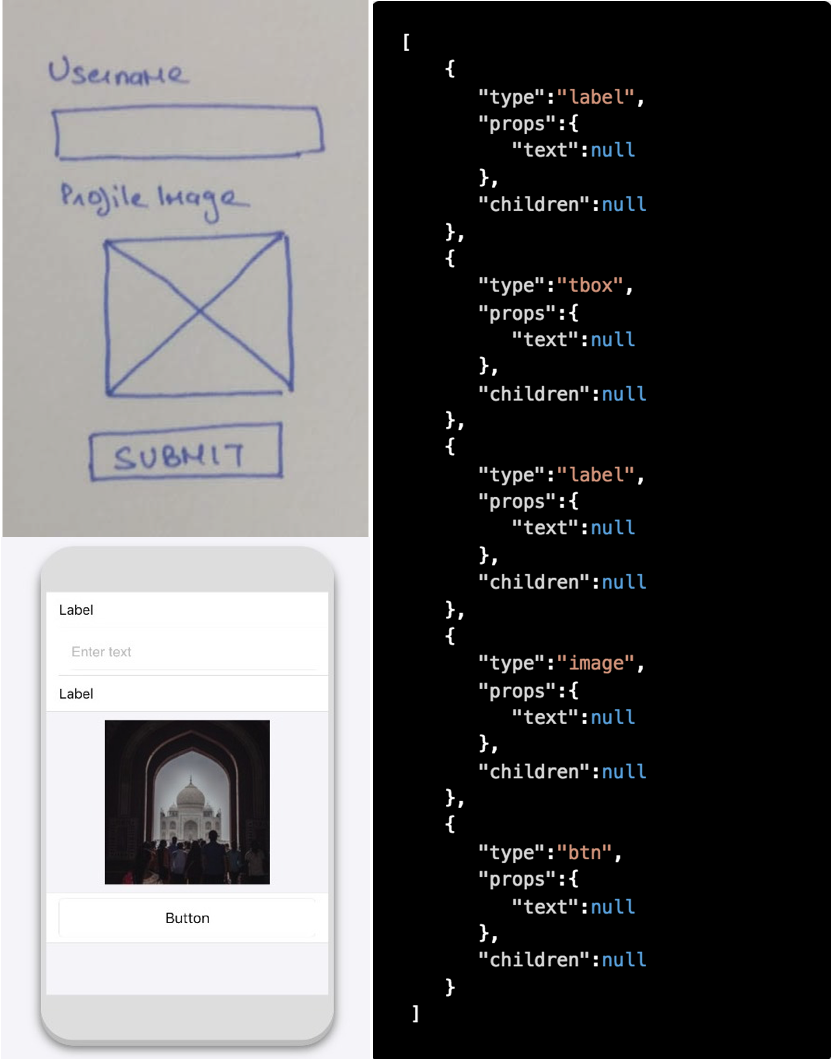}\hfill
    \caption{Input image, Visible coded UI and UI Representation Object}
    \label{output}
\end{figure}

Figure \ref{output} shows the three-step process of conversion of an image into the final coded UI. Initially, a sketch as an image is fed into the application which is then converted into UI representation object. This is a Document Object Model (DOM) based JavaScript Object Notation (JSON) which is then used for the final production of the UI. The model is responsible for generating this UI representation object. Thereafter, any UI parser can be used to generate platform specific UI. In our experiment, we have built a React (a JavaScript UI library) based parser to convert the UI representation into the final prototype application.

\section{Conclusion} \label{sec:conclusion}
The presented work, Sketch2Code, demonstrates that we can train a neural network to identify UI elements in hand-drawn sketches. Performance, which is affected by the training set, can be improved by providing more labelled examples of sketches. While analyzing, we evaluated the model on a set of diverse images. The model learns about the shape of the UI components, and after fixing overlapped components, it gives a final UI representation object allowing the UI parser to create platform specific UI.

\bibliographystyle{IEEEtran}
\bibliography{ieee}

\begin{thebibliography}{10}
\providecommand{\url}[1]{#1}
\csname url@samestyle\endcsname
\providecommand{\newblock}{\relax}
\providecommand{\bibinfo}[2]{#2}
\providecommand{\BIBentrySTDinterwordspacing}{\spaceskip=0pt\relax}
\providecommand{\BIBentryALTinterwordstretchfactor}{4}
\providecommand{\BIBentryALTinterwordspacing}{\spaceskip=\fontdimen2\font plus
\BIBentryALTinterwordstretchfactor\fontdimen3\font minus
  \fontdimen4\font\relax}
\providecommand{\BIBforeignlanguage}[2]{{%
\expandafter\ifx\csname l@#1\endcsname\relax
\typeout{** WARNING: IEEEtran.bst: No hyphenation pattern has been}%
\typeout{** loaded for the language `#1'. Using the pattern for}%
\typeout{** the default language instead.}%
\else
\language=\csname l@#1\endcsname
\fi
#2}}
\providecommand{\BIBdecl}{\relax}
\BIBdecl

\bibitem{Myers:2000:PPF:344949.344959}
\BIBentryALTinterwordspacing
B.~Myers, S.~E. Hudson, R.~Pausch, and R.~Pausch, ``Past, present, and future
  of user interface software tools,'' \emph{ACM Trans. Comput.-Hum. Interact.},
  vol.~7, no.~1, pp. 3--28, Mar. 2000. [Online]. Available:
  \url{http://doi.acm.org/10.1145/344949.344959}
\BIBentrySTDinterwordspacing

\bibitem{eyiokur}
F.~{\.I}. Eyiokur, D.~Yaman, and H.~K. Ekenel, ``Sketch classification with
  deep learning models,'' in \emph{2018 26th Signal Processing and
  Communications Applications Conference (SIU)}.\hskip 1em plus 0.5em minus
  0.4em\relax IEEE, 2018, pp. 1--4.

\bibitem{rian2018fractal}
I.~M. Rian, M.~Sassone, and S.~Asayama, ``From fractal geometry to
  architecture: Designing a grid-shell-like structure using the
  takagi--landsberg surface,'' \emph{Computer-Aided Design}, vol.~98, pp.
  40--53, 2018.

\bibitem{sonmez2018review}
N.~O. S{\"o}nmez, ``A review of the use of examples for automating
  architectural design tasks,'' \emph{Computer-Aided Design}, vol.~96, pp.
  13--30, 2018.

\bibitem{modi2011architecture}
S.~Modi, M.~Tiwari, Y.~Lin, and W.~Zhang, ``On the architecture of a
  human-centered cad agent system,'' \emph{Computer-Aided Design}, vol.~43,
  no.~2, pp. 170--179, 2011.

\bibitem{hablicsek2019algebraic}
M.~Hablicsek, M.~Akbarzadeh, and Y.~Guo, ``Algebraic 3d graphic statics:
  Reciprocal constructions,'' \emph{Computer-Aided Design}, vol. 108, pp.
  30--41, 2019.

\bibitem{sezmfl13}
\BIBentryALTinterwordspacing
P.~Sermanet, D.~Eigen, X.~Zhang, M.~Mathieu, R.~Fergus, and Y.~LeCun,
  ``Overfeat: Integrated recognition, localization and detection using
  convolutional networks,'' 2013. [Online]. Available:
  \url{http://arxiv.org/abs/1312.6229}
\BIBentrySTDinterwordspacing

\bibitem{szegedy}
C.~Szegedy, A.~Toshev, and D.~Erhan, ``Deep neural networks for object
  detection,'' in \emph{Advances in neural information processing systems},
  2013, pp. 2553--2561.

\bibitem{wang}
L.~Wang, W.~Ouyang, X.~Wang, and H.~Lu, ``Visual tracking with fully
  convolutional networks,'' in \emph{Proceedings of the IEEE international
  conference on computer vision}, 2015, pp. 3119--3127.

\bibitem{ddsllf}
J.~Deng, W.~Dong, R.~Socher, L.-J. Li, K.~Li, and L.~Fei-Fei, ``Imagenet: A
  large-scale hierarchical image database,'' in \emph{Proceedings of IEEE
  conference on Computer Vision and Pattern Recognition 2009.}\hskip 1em plus
  0.5em minus 0.4em\relax IEEE, 2009, pp. 248--255.

\bibitem{robust}
C.~Ma, J.-B. Huang, X.~Yang, and M.-H. Yang, ``Robust visual tracking via
  hierarchical convolutional features,'' \emph{IEEE transactions on pattern
  analysis and machine intelligence}, 2018.

\bibitem{erhan}
D.~Erhan, C.~Szegedy, A.~Toshev, and D.~Anguelov, ``Scalable object detection
  using deep neural networks,'' in \emph{The IEEE Conference on Computer Vision
  and Pattern Recognition (CVPR)}, June 2014.

\bibitem{saliency}
S.~He and N.~Pugeault, ``Deep saliency: What is learnt by a deep network about
  saliency?'' \emph{arXiv preprint arXiv:1801.04261}, 2018.

\bibitem{g}
R.~Girshick, ``Fast r-cnn,'' in \emph{Proceedings of the IEEE international
  conference on computer vision}, 2015, pp. 1440--1448.

\bibitem{retnet}
T.-Y. Lin, P.~Goyal, R.~Girshick, K.~He, and P.~Doll{\'a}r, ``Focal loss for
  dense object detection,'' \emph{IEEE transactions on pattern analysis and
  machine intelligence}, 2018.

\bibitem{DBLP:journals/corr/RedmonDGF15}
J.~Redmon, S.~Divvala, R.~Girshick, and A.~Farhadi, ``You only look once:
  Unified, real-time object detection,'' in \emph{Proceedings of the IEEE
  conference on computer vision and pattern recognition}, 2016, pp. 779--788.

\bibitem{DBLP:journals/corr/abs-1804-02767}
\BIBentryALTinterwordspacing
J.~Redmon and A.~Farhadi, ``Yolov3: An incremental improvement,'' \emph{CoRR},
  vol. abs/1804.02767, 2018. [Online]. Available:
  \url{http://arxiv.org/abs/1804.02767}
\BIBentrySTDinterwordspacing

\bibitem{gddm}
R.~Girshick, J.~Donahue, T.~Darrell, and J.~Malik, ``Rich feature hierarchies
  for accurate object detection and semantic segmentation,'' in
  \emph{Proceedings of the IEEE conference on computer vision and pattern
  recognition}, 2014, pp. 580--587.

\bibitem{dean}
T.~Dean, M.~A. Ruzon, M.~Segal, J.~Shlens, S.~Vijayanarasimhan, and J.~Yagnik,
  ``Fast, accurate detection of 100,000 object classes on a single machine,''
  in \emph{Proceedings of the IEEE Conference on Computer Vision and Pattern
  Recognition}, 2013, pp. 1814--1821.

\bibitem{deepcoder}
\BIBentryALTinterwordspacing
M.~Balog, A.~L. Gaunt, M.~Brockschmidt, S.~Nowozin, and D.~Tarlow, ``Deepcoder:
  Learning to write programs,'' \emph{CoRR}, vol. abs/1611.01989, 2016.
  [Online]. Available: \url{http://arxiv.org/abs/1611.01989}
\BIBentrySTDinterwordspacing

\bibitem{b}
T.~Beltramelli, ``pix2code: Generating code from a graphical user interface
  screenshot,'' in \emph{Proceedings of the ACM SIGCHI Symposium on Engineering
  Interactive Computing Systems}.\hskip 1em plus 0.5em minus 0.4em\relax ACM,
  2018, p.~3.

\bibitem{nc15}
T.~A. Nguyen and C.~Csallner, ``Reverse engineering mobile application user
  interfaces with remaui (t),'' in \emph{Automated Software Engineering , 2015
  30th IEEE/ACM International Conference on}.\hskip 1em plus 0.5em minus
  0.4em\relax IEEE, 2015, pp. 248--259.

\bibitem{ocr}
G.~Nagy, T.~A. Nartker, and S.~V. Rice, ``Optical character recognition: An
  illustrated guide to the frontier,'' in \emph{Document Recognition and
  Retrieval VII}, vol. 3967.\hskip 1em plus 0.5em minus 0.4em\relax
  International Society for Optics and Photonics, 1999, pp. 58--70.

\bibitem{canny}
J.~Canny, ``A computational approach to edge detection,'' \emph{IEEE
  Transactions on Pattern Analysis and Machine Intelligence}, vol.~8, no.~6,
  pp. 679--698, Nov 1986.

\bibitem{edgedil}
N.~Sengar and D.~Kapoor, ``Edge detection by combination of morphological
  operators with different edge detection operators,'' \emph{International
  Journal of Information \& Computation Technology}, vol.~4, no.~11, pp.
  1051--1056, 2014.

\bibitem{zhiqiang2017review}
W.~Zhiqiang and L.~Jun, ``A review of object detection based on convolutional
  neural network,'' in \emph{Control Conference , 2017 36th Chinese}.\hskip 1em
  plus 0.5em minus 0.4em\relax IEEE, 2017, pp. 11\,104--11\,109.

\bibitem{lstm}
T.~N. Sainath, O.~Vinyals, A.~Senior, and H.~Sak, ``Convolutional, long
  short-term memory, fully connected deep neural networks,'' in \emph{2015 IEEE
  International Conference on Acoustics, Speech and Signal Processing
  (ICASSP)}.\hskip 1em plus 0.5em minus 0.4em\relax IEEE, 2015, pp. 4580--4584.

\bibitem{fpn}
T.-Y. Lin, P.~Doll{\'a}r, R.~Girshick, K.~He, B.~Hariharan, and S.~Belongie,
  ``Feature pyramid networks for object detection,'' in \emph{Proceedings of
  the IEEE Conference on Computer Vision and Pattern Recognition}, 2017, pp.
  2117--2125.

\bibitem{websocket}
\BIBentryALTinterwordspacing
I.~Fette and A.~Melnikov, ``The websocket protocol,'' 2011. [Online].
  Available: \url{https://tools.ietf.org/html/rfc6455}
\BIBentrySTDinterwordspacing

\bibitem{hzrs}
K.~He, X.~Zhang, S.~Ren, and J.~Sun, ``Deep residual learning for image
  recognition,'' in \emph{Proceedings of the IEEE conference on computer vision
  and pattern recognition}, 2016, pp. 770--778.

\bibitem{ksh12}
A.~Krizhevsky, I.~Sutskever, and G.~E. Hinton, ``Imagenet classification with
  deep convolutional neural networks,'' in \emph{Advances in neural information
  processing systems}, 2012, pp. 1097--1105.

\end{thebibliography}
\end{document}